\documentclass[10pt,twocolumn,letterpaper]{article}

\usepackage[pagenumbers]{wacv} 

\usepackage{graphicx}
\usepackage{amsmath}
\usepackage{amssymb}
\usepackage{booktabs}
\usepackage{soul}

\usepackage[pagebackref,breaklinks,colorlinks]{hyperref}

\usepackage[capitalize]{cleveref}
\crefname{section}{Sec.}{Secs.}
\Crefname{section}{Section}{Sections}
\Crefname{table}{Table}{Tables}
\crefname{table}{Tab.}{Tabs.}

\def\wacvPaperID{2602} 
\def\confName{WACV}
\def\confYear{2025}

\newcommand{\mname}{MCD}
\usepackage{multirow}
\usepackage[dvipsnames]{xcolor}
\usepackage{verbatim}
\usepackage{float}
\usepackage{listings}
\begin{document}

\title{Generation of Complex 3D Human Motion by Temporal and Spatial Composition of Diffusion Models}

\author{Lorenzo Mandelli\\
Florence University\\
{\tt\small lorenzo.mandelli@unifi.it}
\and
Stefano Berretti\\
Florence University\\
{\tt\small
stefano.berretti@unifi.it}
}
\twocolumn[{
\maketitle
\vspace{-0.5cm}
\begin{center}
\includegraphics[width=\textwidth]{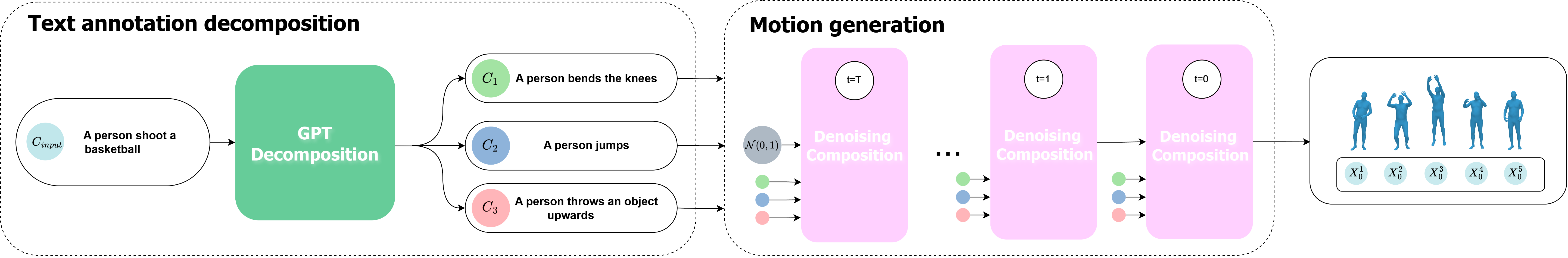}
\captionof{figure}{The proposed 3D human motion generation pipeline during the inference phase. It is divided into two stages: first, an input textual annotation, which was not seen during the training phase, is divided into simple actions through a GPT \textit{decomposition} module. Then, each simple action contributes at each step of the denoising diffusion process to generate the final motion.}
\label{fig:full_pipeline}
\end{center}
}]

\begin{abstract}
In this paper, we address the challenge of generating realistic 3D human motions for action classes that were never seen during the training phase. Our approach involves decomposing complex actions into simpler movements, specifically those observed during training, by leveraging the knowledge of human motion contained in GPTs models. These simpler movements are then combined into a single, realistic animation using the properties of diffusion models.
Our claim is that this decomposition and subsequent recombination of simple movements can synthesize an animation that accurately represents the complex input action. This method operates during the inference phase and can be integrated with any pre-trained diffusion model, enabling the synthesis of motion classes not present in the training data.
We evaluate our method by dividing two benchmark human motion datasets into basic and complex actions, and then compare its performance against the state-of-the-art.
\end{abstract}

\vspace{-0.3cm}
\section{Introduction}
\label{sec:intro}
Driven by applications in video games, the movie industry, and entertainment, recent research has shown significant advancements in learning to generate 3D human motion~\cite{HumanML3D, momask, singlemotion}. One of the most significant tasks is text-conditioned generation that aims to enable non-expert users to create 3D animations intuitively by simply specifying an input prompt. Several solutions have been proposed for this task~\cite{human_motion_diffusion_model, latent_diffusion, tm2t}.

While these methods are a promising step forward in developing interfaces for animation generation, their ability to create movements is heavily dependent on the motions present in the dataset. To achieve movements of a class not present in the training set, it is necessary to record a person performing the action using appropriate sensors, add the motion to the dataset, and then retrain a generative model with the additional data. 

Our intuition is that any unknown action can be decomposed into a set of simple and known actions that, when combined, replicate the original movement without the need to collect new data. For instance, \textit{``A person shoots a basketball''} can be broken down into the sequence of actions \textit{``bend knees''}, \textit{``jump''}, and \textit{``throw an object upwards''}. If our dataset does not contain any action related to \textit{``basketball''} but only basic actions like \textit{``bending''}, \textit{``jumping,''} and \textit{``throwing an object,''} we can still handle the case through \textit{decomposition}. This approach mirrors how humans learn a new movement by breaking it down into familiar sub-movements and then combining them.

Specifically, the literature distinguishes between two types of possible compositions of human motions: \textbf{Temporal}~\cite{teach}: two movements executed one after the other, such as \textit{``bend knees''} and then \textit{``jump''};  \textbf{Spatial}~\cite{sinc}: two movements occurring simultaneously, such as \textit{``jump''} while \textit{``throwing an object''}.

Motivated by these considerations, in this paper we propose our method called \mname~(Motion Composition Diffusion) for generating actions unseen during the training phase. \mname~takes a textual annotation describing the movement to be generated and its duration and decomposes it through GPT into one or more known sub-movements present in the training set, each characterized by a start and an end time.

During inference, at every step of diffusion denoising, our method processes each sub-movement individually and then combines them based on their respective start and end times. Our method offers a general solution to both temporal and spatial decomposition, enabling the execution of both without distinction. Operating during the inference phase, our method can be integrated with any pre-trained motion diffusion model to synthesize motions that fall outside the distribution of the training set.

To prove the effectiveness of our approach, we divide the datasets into \textit{basic} actions and \textit{complex} actions, train a diffusion model on the basic actions, and test it on the complex actions. We then compare our method with the implementation without decomposition but conditioned only by text and with other composition strategies in the literature. 
Additionally, we test our composition method on the task of synthesizing animations by leveraging the multiple textual annotations for each elements available in the datasets. In this setting, we achieve better results than text-conditioned generation alone.

In summary, our main contributions are as follows:
\begin{itemize}
\itemsep0em 
    \item We propose a generation method capable of producing human motion belonging to classes not present in the train set distribution;
    \item A new general, constraint-free method for the spatial and temporal composition of human motion;
    \item We demonstrate the effectiveness of our approach by dividing well-known datasets into splits of simple, known actions and complex, unseen actions;
    \item We demonstrate that by utilizing multiple textual annotations simultaneously for a single dataset element, we achieve better results compared to traditional text-conditioned generation.
\end{itemize}

\section{Related work}
\label{sec:related_work}
\textbf{Human motion generation.} Significant progress has been made in the generation of 3D human movements. 
Architecturally, various approaches have been explored, including the use of VAEs~\cite{VAE_0, temos}, GANs~\cite{Gan_0, Gan_1}, Normalizing Flows~\cite{Normalizing_flow_0}, and more recently, Diffusion Models~\cite{latent_diffusion, human_motion_diffusion_model}. 
Furthermore, both unconditional generation~\cite{uncoditioned} and conditional generation based on actions~\cite{Conditioned_action_0}, speech~\cite{conditioned_speech_0, conditioned_speech_1}, text~\cite{conditioned_text_0, conditioned_text_1, conditioned_text_2, conditioned_text_3, conditioned_text_4}, music~\cite{conditioned_music_0, conditioned_music_1, conditioned_music_2} have been studied.

Our focus is on text-conditioned generation using Diffusion Models, with a particular emphasis on a novel approach to generate unseen actions by combining knowledge from known actions.

\textbf{Guidance in Diffusion Models}.
Several studies have explored the capabilities of diffusion models to be conditioned  in a manner that enables precise control over the generated output based on specific input.
In~\cite{guidance_classifier}, the authors proposed classifier guidance, a technique to enhance the sample quality of a diffusion model using an additional trained classifier. 
In~\cite{guidance_classifier_free}, authors jointly trained a conditional and an unconditional diffusion model, combining the resulting conditional and unconditional score estimates to achieve a balance between sample quality. During the sampling procedure they perform a linear combination of the conditional and the unconditional terms by using:
\begin{equation}
\tilde{\epsilon}_\theta(z_\lambda, c) = (1+w)\epsilon_\theta(z_\lambda, c) - w\epsilon_{\theta}(z_\lambda) \label{eq:classifier_free_score} ,
\end{equation}

\noindent 
where $c$ represents the input condition such as text or action, and $w$ is a parameter that controls the strength of the classifier guidance. 

\textbf{Composable Generation}. Several works have explored the capabilities of VAEs and diffusion models to combine multiple pieces of information for composite generation in both image and 3D generation domains. 
In~\cite{teach}, in order to solve \textit{temporal composition}, an autoregressive approach is used to temporally concatenate individually generated animations in a non-autoregressive manner, conditioning each subsequent motion on the previous one. 
In~\cite{sinc}, \textit{spatial composition} is addressed by using GPT-3 to determine, which body parts are necessary for specific actions and using this information to crop and combine individual actions that occur simultaneously.
In~\cite{Composable_1} to address the issues of attribute leakage and entity leakage, the generative process is guided by a box predictor and a mask control mechanism that leverages the cross- and self-attention maps of the predicted boxes. 
In~\cite{Composable_0}, the similarity between diffusion models and Energy-Based Models (EBM) is examined, enabling the composition of the former through AND and OR logic functions. 
In~\cite{diffCollage}, a compositional strategy based on diffusion models is employed to generate large content by leveraging diffusion models trained on generating small portions of the overall images.

\begin{figure*}[h]
    \centering
    \includegraphics[width=0.7\linewidth]{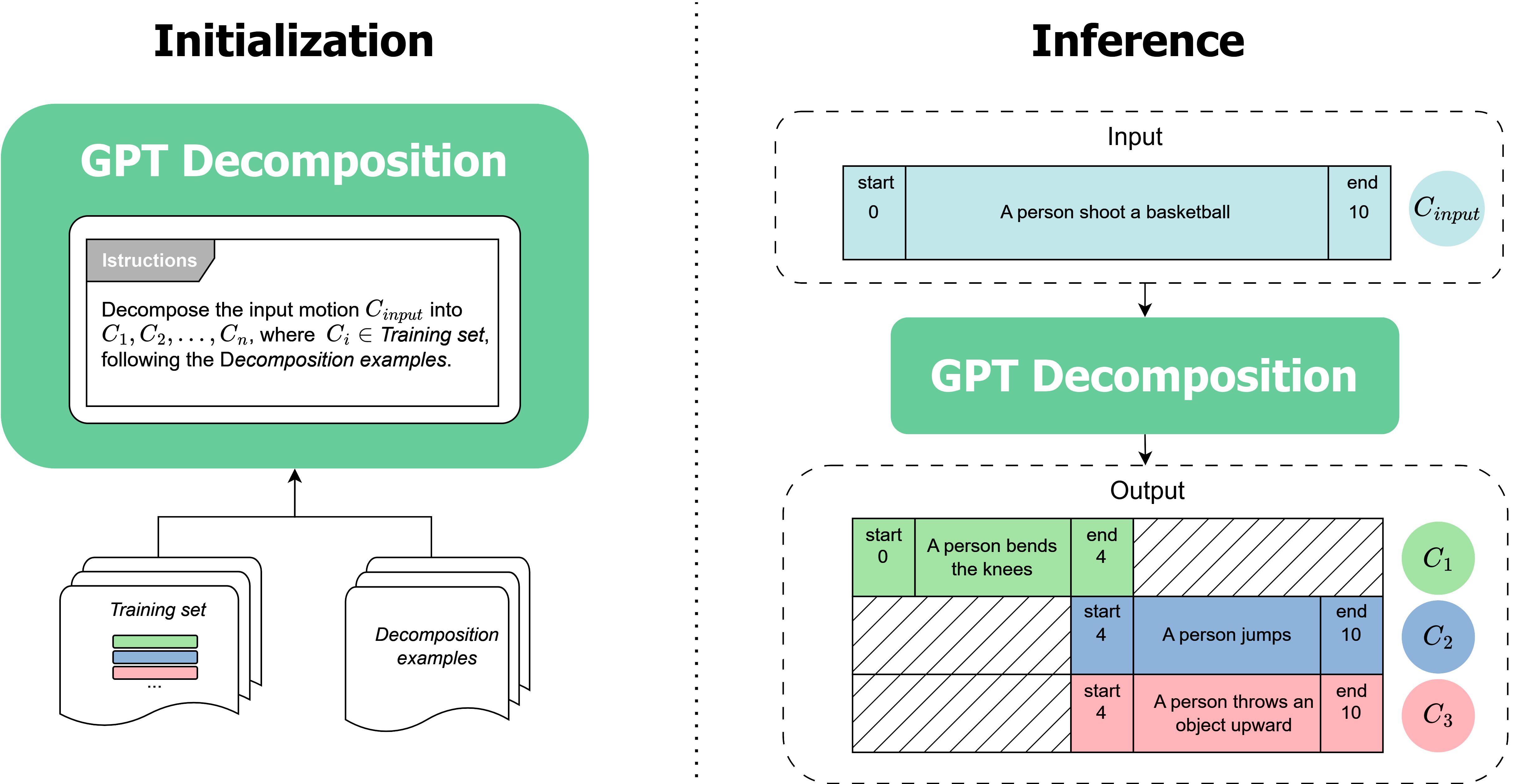}
    \caption{\textbf{(Left)}  The GPT module for annotation decomposition is initialized using the information provided in the instructions, decomposition examples, and known actions available in the training set. Its goal is to decompose any input motion annotation into sub-movement annotations present in the training set, following the provided examples, and to assign temporal boundaries to each of them. \textbf{(Right)} An input motion annotation, \( C_{\text{input}} \), characterized by a textual description (``A person shoot a basketball'') and a duration interval (start at $time = 0$, end at $time = 10$), is decomposed into a series of sub-movement annotations \( C_1, C_2, \ldots, C_n \) through a call based on OpenAI's GPT. We note that generated sub-movements can connect either temporally (the green with the blue and red ones) or spatially (the blue and red sub-movements).}
   \label{fig:gtp-decomposition}
   \vspace{-0.2cm}
\end{figure*}

Drawing inspiration from the advancements in GPT models and the capabilities of diffusion models to be combined, we tackle the task of generating actions that were never seen during the training phase by decomposing them into known actions that can be combined. We leverage the impressive abilities recently demonstrated by GPT-3 and GPT-4, and following the approach in~\cite{Composable_0}, we treat concurrent actions to be combined spatially as an \textbf{AND} condition. The most closely related work to ours is~\cite{stmc}, that we compared with in the experimental evaluation. We differentiate ourselves from this approach in three key ways: \emph{(i)} our focus on the task of generating unseen actions, \emph{(ii)} the inclusion of a decomposition phase, which is unique to our work, and \emph{(iii)} our strategy for combining human movements, which offers a more general approach to the combination of diffusion models. Furthermore, unlike this previous work, our method does not require assigning specific body parts to each input movement in order to crop them together during the denoising phase.

\section{Proposed Method}
In the following, in Section~\ref{sect:diffusion-model}, we first summarize the essential concepts of diffusion models and the related notation; then, we rely on these to introduce our approach in Section~\ref{sect:our-method}. 
\begin{figure*}[h]
    \centering 
    \includegraphics[width=0.9\linewidth]{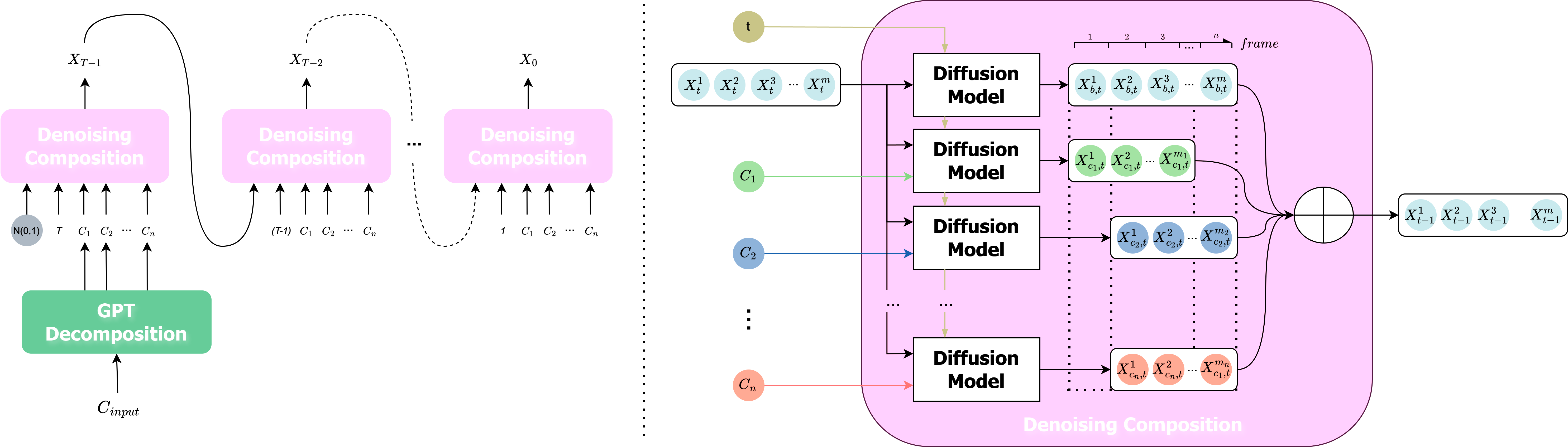}
    \caption{\textbf{(Left)} Sampling algorithm based on input decomposition: the unknown input motion, $C_{input}$, is decomposed into $n$ submovements [\(c_1, c_2, \ldots, c_n\)], which collectively condition the generative process to reconstruct the original motion. \textbf{(Right)} Denoising composition over $n$ submovements: at each denoising step, each $i$-submovement is generated based on the corresponding $C_i$. Then, all the generated motions are merged together according to their respective duration intervals into a single motion.}
   \label{fig:composition-pipeline}
   \vspace{-0.3cm}
\end{figure*}

\subsection{Diffusion model} \label{sect:diffusion-model}
Denoising Diffusion Probabilistic Models (DDPMs) are a class of generative models, where the generation process is formulated as a denoising task. Beginning with sampled noise, the diffusion model performs $T$ denoising steps to produce a clean motion $x_0$. The process generates a series of intermediate motions with progressively decreasing noise levels, denoted as $x_T, x_{T-1}, \ldots, x_0$, where $x_T$ is sampled from a Gaussian prior and $x_0$ is the final output motion. DDPMs construct a forward diffusion process by incrementally adding Gaussian noise to the ground truth. The model then learns to reverse this noise corruption process. Both the forward process $q(x_t | x_{t-1})$ and the reverse process $p_\theta(x_{t-1} | x_t)$ are modeled as products of Markov transition probabilities:
\begin{align}
q(x_{0:T}) = q(x_0) \prod_{t=1}^{T} q(x_t | x_{t-1}) ,
\\
p_\theta(x_{0:T}) = p(x_T) \prod_{t=1}^{T} p_\theta(x_{t-1} | x_t) ,
\end{align}

\noindent
where $q(x_0)$ is the real data distribution and $p(x_T)$ is a standard Gaussian prior. The generative process $p_\theta(x_{t-1} | x_t)$ is trained to generate realistic motions by approximating the reverse process through variational inference. Each step of the generative process is a Gaussian distribution $\mathcal{N}$ with a learned mean $\mu_\theta(x_t, t)$ and covariance matrix $\sigma_t^2 \mathbf{I}$, where $\mathbf{I}$ is the identity matrix.
The mean $\mu_\theta(x_t, t)$ is represented by a perturbation $\epsilon_\theta(x_t, t)$ to the noisy motion $x_t$. The objective is to gradually remove the noise by predicting a less noisy motion at timestep $x_{t-1}$ given a noisy motion $x_t$. To generate clean animations, we sample $x_{t-1}$ from $t = T$ to $t = 0$ using the parameterized marginal distribution $p_\theta(x_{t-1} | x_t)$, with an individual step: 
\begin{equation}
p_\theta(x_{t-1} | x_t) = \mathcal{N}(x_{t-1}; \mu_\theta(x_t, t), \sigma_t^2 \mathbf{I}) . 
\end{equation}

Over multiple iterations, the generated 3D human motion becomes more and more realistic and smooth.

In our experiments, instead of predicting $\epsilon_t$, we follow~\cite{predict_x0} and predict the signal itself $\hat{x}_0$, as we found this strategy provides better results for our case. Furthermore, following~\cite{stmc}, we utilize a model based on~\cite{human_motion_diffusion_model} that predicts an SMPL body representation instead of body joints. 

\subsection{\mname} \label{sect:our-method}
\mname~enables the generation of unseen actions during the inference phase and operates in two stages as shown in~\cref{fig:full_pipeline}:
(1) The textual description of an unknown input movement is decomposed using GPT into multiple known basic actions, each with associated time intervals. For simplicity, we will refer to these decomposed elements as \textit{``sub-movements''} from now on; 
(2) At each step of the denoising diffusion process, each sub-movement motion is generated, then all are combined into a single movement, taking into account their respective time intervals. 

More details for these two phases are given below. 

\vspace{-0.4cm}
\subsubsection{Unknown action decomposition}\label{sec:gpt_decomposition}
As shown in~\cref{fig:gtp-decomposition}, we use GPT-4o-mini \cite{gpt4} to decompose an input action into basic actions. 
Specifically, we utilize GPT-assistance to exploit its ability to retrieve elements from a know text file.
Finally, to achieve the decomposition task, we compose a GPT prompt based on three components: \emph{(i)} the instructions, which provide guidelines for the decomposition task; \emph{(ii)} an example file containing 50 handcrafted examples to guide GPT in producing the desired output; and \emph{(iii)} a file of known actions, which includes the text of actions available during the training phase. 
The instructions direct GPT to break down the input actions into one or more sub-actions using those listed in the provided file, while following the logic demonstrated in the examples. For each sub-movement, the start and end times are required. The goal is for the set of all generated sub-movements, when recombined according to their temporal intervals, to be equivalent to the input movement. 
An example of such a decomposition is shown in~\cref{fig:gtp-decomposition}.

\vspace{-0.2cm}
\subsubsection{Composition of submovements}
We aim to combine $n$ annotations of sub-movements, each characterized by a text description, start time and end time. At each denoising step, we generate an unconditional \textit{base} movement $x_{\text{b}}$ with a duration matching the input requirements, and $n$ sub-movements [$x_1,..,x_n$] for each decomposed $i$-elements with durations specified by their respective start and time annotations $s_i$ and $e_i$.
Finally, at each denoising step the $N+1$ generated movements  are combined together into one using:
\begin{equation}
\begin{split} 
    \label{eq:composition}
    x_{t-1} = x_b(t) +  w \sum_{i=1}^{n} \bigl( x_i(t | {c}_i)_{[s_i, e_i]} - x_{b}(t)_{[s_i, e_i]} \bigl)
\end{split}
\end{equation}

\noindent
where $w$ is the hyper-parameter that controls the "strength" of the spatial combination, $c_i$ is  the text annotation that conditions the output $x_i$, and $[s_i, e_i]$ indicates that we take the motion between the time boundaries $s_i$ and $e_i$ of sub-movement $i$. Figure~\ref{fig:composition-pipeline} shows an overview of our sub-movements composition method at inference time. 

Intuitively, $x_{\text{c}}$ provides a foundational structure for the overall movement, upon which the various $i$-sub-movements are applied within their respective time intervals. The progressive denoising process refines the resulting motion, bringing it as close as possible to the data distribution, ensuring that the generated movement remains smooth despite the different contributions.

As argued by~\cite{Composable_0}, the sum operation in~\eqref{eq:composition} for annotations belonging in the same time interval corresponds to an AND condition. This means that it simultaneously satisfies two textual conditions, allowing for spatial composition. By summing the sub-movements generated within their respective time intervals, we can specify when and for how long the actions occur, thus enabling temporal composition. In this way, both spatial and temporal compositions are achieved using a single equation.

Unlike the approach in~\cite{stmc}, our method is agnostic to spatial overlap and does not require the following constraints: \emph{(i)} that each sub-movement specifies the primary body part involved, such as \textit{``legs''},\textit{``left arm''},\textit{``right arm''}, \textit{etc.}, \emph{(ii)} that no single body part is assigned to two different sub-movements within the same time interval. 
In contrast, our method allows for the spatial combination of sub-movements that involve the same body part. For example, ``jumping'' and ``kicking,'' both of which require the ``legs,'' can still be combined into a single movement like a ``jumping kick``. 
The approach in~\cite{stmc} builds on the composition equation from~\cite{diffCollage}, which is a specific case of the AND formula found in~\cite{Composable_0}, where $w=1$. 
Our method, however, draws inspiration from a more generalized version of this equation, where $w$ can assume any value. 
Thus, we offer a more general and direct solution to the problem of human motion composition using diffusion models.

\vspace{-0.2cm}
\section{Experiments}

\subsection{Datasets}
\textbf{HumanML3D}~\cite{HumanML3D} is a 3D human motion-language dataset that combines the HumanAct12~\cite{Guo_2020} and the Amass~\cite{AMASS} motion capture datasets. This dataset comprises 44,970 text annotations corresponding to 14,616 motions. It includes both elementary actions such as ``bending'' and ``walking,'' as well as specific sport performances like karate, swimming, and tennis.

\textbf{KitML}~\cite{KIT} is a text-motion dataset that provides textual descriptions for a subset of the KIT WholeBody Human Motion Database~\cite{KITWholeBody} and of the CMU Graphics Lab Motion Capture Database~\cite{CMU}. This dataset consists of 3,911 motion sequences with 6,353 text descriptions.

\vspace{-0.2cm}
\subsubsection{Splits between \textit{base} and \textit{complex} actions} \label{splits}
To test the ability of our method to generate motions not present in the training phase, we split the datasets into \textit{basic} actions, to be used during training, and \textit{complex} actions, to be generated during testing. The mentioned datasets are characterized by textual annotations and do not natively include class divisions. We then define a movement as \textit{complex} if it represents a sport or, in general, belongs to one of the following categories: \textbf{ball sports} such as baseball, tennis, golf;  \textbf{martial arts} such as kickbox, karate, boxe; \textbf{dances} such as salsa, cha cha, walzer; \textbf{musical instrument} such as drums, guitar, piano.

We classify any element in the dataset as \textit{complex} if it contains at least one of the keywords above (or a derived form). We classify the remaining elements as \textit{basic}. Examples of \textit{basic} elements include ``jumping,'' ``walking,'' ``throwing an object,'' \etc.

Using this approach and augmenting the datasets with mirrored elements (\eg, changing the term ``left arm'' to ``right arm'' and mirroring the motion accordingly), we obtain a split for the HumanML3D dataset characterized by over 25,000 elements for the training set, more than 1,700 for the validation set, and over 2,900 for the test set. Similarly, for the KitML dataset, we obtain over 5,000 elements for the training set, more than 300 for the validation set, and over 400 for the test set.

\subsection{Metrics}
Following~\cite{stmc} we evaluate our approach using the TMR model ~\cite{tmr}, which operates similarly to CLIP~\cite{clip} for images and texts, by providing a joint embedding space to determine the similarity between text and motion.
Using this model, we can calculate the similarity between movements and between movement and text. 
We then add the TMRScore metrics: 
%
M2M, calculated between ground truth movements and generated motions; and
M2T, calculated between the generated motions (or the ground truth motions for baseline) and the ground truth texts.

Furthermore, we employ the metrics: 
%
FID that measures the dissimilarity between the generated and ground truth distributions;
\textit{Transition} distance, that measures the speed of the animation by calculating the mean Euclidean distance (in cm) between the poses of every two consecutive frames;
$R$-\textit{precision} (as in ~\cite{diverse_and_natural}) that measures the relevance of the generated motions to the input prompts as the frequency of the correct text prompt being in the top-1 (R@1), top-3 (R@3) and top-10 (R@10) retrieved texts from the entire set of annotation texts.

\begin{table*}[h!]
\centering
\footnotesize
\caption{Comparison between normal text-conditioned generation, the STMC composition approach, and our method (GPT Decomposition + Motion Composition)   on the HumanML3D and KITML datasets over the splits of \cref{splits}. 
}
\begin{tabular}{llccccccc}
\toprule
\textbf{Dataset} & \textbf{Experiment} & \textbf{R1 $\uparrow$} & \textbf{R3 $\uparrow$} & \textbf{R10 $\uparrow$} & \textbf{M2T$\uparrow$} & \textbf{M2M$\uparrow$} & \textbf{FID $\downarrow$} & \textbf{Trans $\rightarrow$} \\
\midrule
\multirow{4}{*}{\textbf{HumanML3D}} 
    & GT          & 34.087 & 61.629 & 70.527 & 0.864 & 1.0  & 0.0  & 1.785 \\
    \cmidrule(lr){2-9}
    & Text-conditioned     & 0.659  & 1.789  & 4.896  & 0.554 & 0.542& 0.59 & 0.432 \\
    & STMC & 0.565 & 2.260 &  5.461 & 0.547 & 0.543 & 0.56 & 0.558 \\
    & Ours (De+Com)               & \textbf{1.177}  & \textbf{3.013}  & \textbf{6.450}  & \textbf{0.593} & \textbf{0.589} &\textbf{0.35 } & \textbf{1.696} \\
\midrule
\multirow{4}{*}{\textbf{KITML}} 
    & GT &  11.25 & 29.00 & 46.25 & 0.881 & 1.0   & 0.0   & 1.227 \\
    \cmidrule(lr){2-9}
    & Text-conditioned & 1.250  & 3.750  & 9.500   & 0.569 & 0.571 & 0.750  & 0.795 \\
    & STMC & 2.500   & 4.250  & 10.50  & 0.595 & 0.582 & 0.672 & 0.762\\
    & Ours (De+Com)  & \textbf{2.750}  & \textbf{6.500} & \textbf{12.50} & \textbf{0.625} & \textbf{0.621} & \textbf{0.519} & \textbf{0.964} \\\bottomrule
\end{tabular}
\label{tab:text-vs-STMC-vs-our_KITML_Humanml}
\end{table*}

\begin{figure*}[t]
    \centering
    \includegraphics[width=\linewidth]{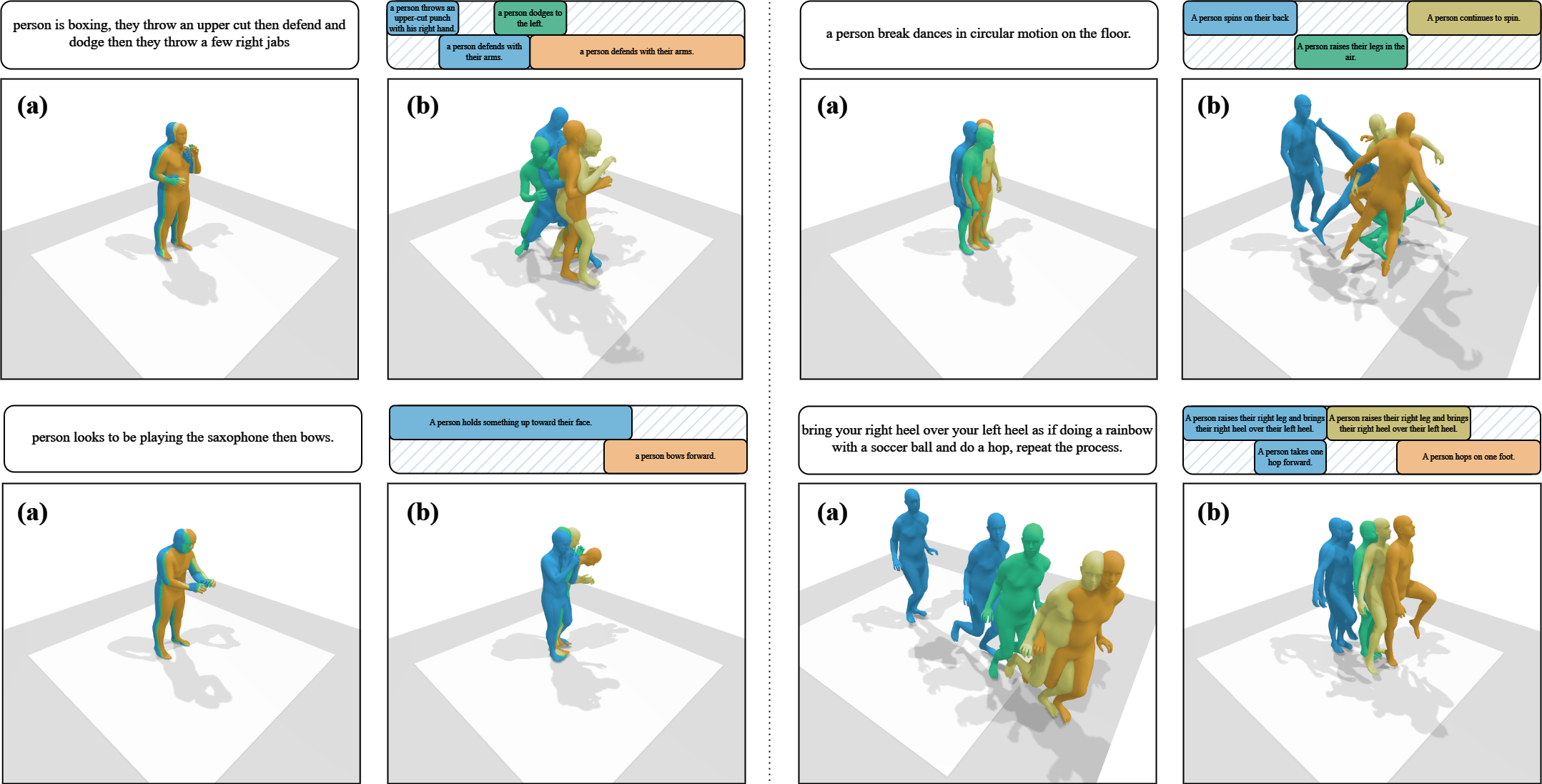}
    \caption{Comparison between text-only conditioned generation \textbf{(a)} and our decomposition-based approach \textbf{(b)}: Text-only conditioning tends to produce stationary animations and fails to execute any meaningful actions if the textual annotations conditioning the movement do not fall within the training distribution. In contrast, by decomposing movements into sub-movements, we are able to successfully generate even complex actions.}
   \label{fig:renders-2}
   \vspace{-0.3cm}
\end{figure*}

\subsection{Evaluation}
\subsubsection{Single text vs. composition}
We trained a diffusion model over the splits described in \cref{splits}, and used it to compare text-conditioned generation without decomposition with our approach. 
The obtained results are shown in~\cref{tab:text-vs-STMC-vs-our_KITML_Humanml}.
For both datasets, we observe higher retrieval scores (R1, R2, R10), as well as greater alignment between the embeddings of the generated data and the ground truth motions and texts (M2M, M2T). Additionally, the FID values between the generated and original distributions also demonstrate improved performance. This indicates that the actions generated not only semantically align with the initial textual conditions but that the distribution of the generated data closely resembles that of the original dataset. 
We obtain a \textit{Transition} distance that is higher compared to the original dataset, but closer to the reference values than the text-only conditioned generation. The lower values for the latter are largely due to the fact that many animations generated without decomposition remain almost static when the textual annotation they were conditioned on is not present during training. This is illustrated in~\cref{fig:renders-2}, where many text-conditioned animations fail to perform any meaningful action and instead remain stationary.
The higher \textit{Transition} values in our approach stem from the 
contributions of sub-movements. During generation, the model attempts to incorporate all sub-movements within their respective time frames, and this multitude of constraints leads to 
faster animations.

\begin{table*}[h!]
\centering
\footnotesize
\caption{Comparison between the text-conditioned generation and our multi-annotations approach of \cref{sec:multiannotations} (Motion Composition) on the HumanML3D and KITML datasets. 
}
\begin{tabular}{llccccccc}
\toprule
\textbf{Dataset} & \textbf{Experiment} & \textbf{R@1 $\uparrow$} & \textbf{R@3 $\uparrow$} & \textbf{R@10 $\uparrow$} & \textbf{M2T$\uparrow$} & \textbf{M2M$\uparrow$} & \textbf{FID $\downarrow$} & \textbf{Trans $\rightarrow$} \\
\midrule
\multirow{3}{*}{\textbf{HumanML3D}} 
    & GT & 6.259 & 15.487 & 33.548 & 0.762 & 1.0 & 0.0 & 1.368 \\
    \cmidrule(lr){2-9}  
    & Text-conditioned & 1.781   & 5.022   & 12.618  & 0.662  & 0.684  & 0.177  & 0.846 \\
    & Ours (Com) & \textbf{4.503}   & \textbf{10.168}  & \textbf{23.454}  & \textbf{0.736}  & \textbf{0.743}  & \textbf{0.168}  & \textbf{1.144} \\
\midrule
\multirow{3}{*}{\textbf{KITML}} 
    & GT             & 7.888  & 19.847  & 41.858  & 0.803  & 1.0   & 0.0   & 1.534 \\
    \cmidrule(lr){2-9}  
    & Text-conditioned & 5.598  & 14.631  & 32.443  & 0.753  & \textbf{0.740}   & \textbf{0.182}  & 1.058  \\
    & Ours (Com)  &  \textbf{7.761}  & \textbf{18.193}  & \textbf{35.623}  & \textbf{0.755}  & 0.731  & 0.192  & \textbf{1.069}  \\
\bottomrule
\end{tabular}
\label{tab:multiannotations_humanml_and_kitml}
\end{table*}

\begin{table*}[h!]
\centering
\footnotesize
\caption{Ablation study of the $w$ parameter over the split in \cref{splits} of the datasets HumanML3D and KITML. 
}
\begin{tabular}{llccccccc}
\toprule
\textbf{Dataset} & \textbf{Experiment} & \textbf{R@1 $\uparrow$} & \textbf{R@3 $\uparrow$} & \textbf{R@10 $\uparrow$} & \textbf{M2T$\uparrow$} & \textbf{M2M$\uparrow$} & \textbf{FID $\downarrow$} & \textbf{Trans $\rightarrow$} \\
\midrule
\multirow{5}{*}{\textbf{HumanML3D}} 
    & $w=1$     & 0.235   & 0.942   & 2.684   & 0.525 & 0.516 & 0.678 & 0.289 \\
   & $w=3$ & 1.036 & 2.307 & 6.215 & 0.576 & 0.572 & 0.400 & 0.972 \\
    & \(w=\underline{5}\)     & \textbf{1.177}   & \textbf{3.013}   & 6.450    & 0.593 & 0.589 & \textbf{0.350}  & \textbf{1.696} \\
  & $w=10$    & \textbf{1.177}   & \textbf{3.013}   & \textbf{6.497}   & 0.600 & \textbf{0.598} & \textbf{0.350} & 2.867 \\
  & $w=15$ & 0.895 & 2.684 & 5.367 & \textbf{0.601} & 0.597 & 0.365 & 3.494 \\
\midrule
\multirow{5}{*}{\textbf{KITML}} 
    & $w=1$   & 1.25  & 3.75  & 9.00   & 0.557 & 0.546 & 0.799 & 0.532 \\
    & $w=5$   & 1.75  & 4.50   & \textbf{13.0}  & 0.612 & 0.600   & 0.606 & 0.776 \\
    & \(w=\underline{10}\)  & \textbf{2.75}  & \textbf{6.50}   & 12.5  & \textbf{0.625} & \textbf{0.621} & 0.519 & 0.964 \\
    & $w=15$  & 2.25  & 5.00   & 10.75 & 0.617 & 0.618 & \textbf{0.511} & \textbf{1.141} \\
    & $w=25$  & 1.25  & 2.75  & 6.75  & 0.609 & 0.620  & 0.514 & 1.770  \\
\bottomrule
\end{tabular}
\label{tab:ablation_w_kitml_humanml_splitme}
\end{table*}

\vspace{-0.2cm}
\subsubsection{STMC vs. composition}
For the comparison with~\cite{stmc}, we decompose the textual annotations using a prompt for our method and one for their approach. Utilizing two separate decompositions, rather than a single one, is necessary due to differing constraints. 
In~\cite{stmc}, each decomposition requires a list of involved body parts (\eg, left arm, right arm, legs, spine, and head), it is not allowed for the same body part to appear in two overlapping time intervals, and there should be no time intervals within the movement that are not assigned to any sub-movement. These constraints arise because, during the diffusion step, the movements are generated and cropped into a single animation based on the indicated body part; having two movements involving the same body part would create ambiguity in the cropping process. 

In contrast, our method is agnostic to specific body parts and only requires the temporal specifications of the sub-movements. Conceptually, this allows us to combine actions involving the same body parts. For example, actions like ``jumping'' and ``kicking,'' which both involve the legs, can be combined to create a ``flying kick,'' something not feasible with the approach in~\cite{stmc}.

Both prompts necessary for the comparison consist of a set of instructions and a set of examples to provide guidelines to GPT. To ensure a fair comparison, the same set of examples is reformulated according to the constraints of the method used for each prompt. The prompts used are provided in the \textit{Supplementary Material}.

Results in~\cref{tab:text-vs-STMC-vs-our_KITML_Humanml} show that our method outperforms STMC in terms of retrieval scores (R1, R2, R10), similarity to ground truth embeddings (M2M, M2T), Transition distance and FID values. The differences in performance between the two methods are attributed not only to the compositional approach itself but also to the differing constraints of the GPT Decomposition module \cref{sec:gpt_decomposition} for each method. STMC requires many more constraints, which limits the current GPT's ability to accurately decompose an action, resulting in worst outcomes.


\begin{figure*}[ht]
    \centering    \includegraphics[width=\linewidth]{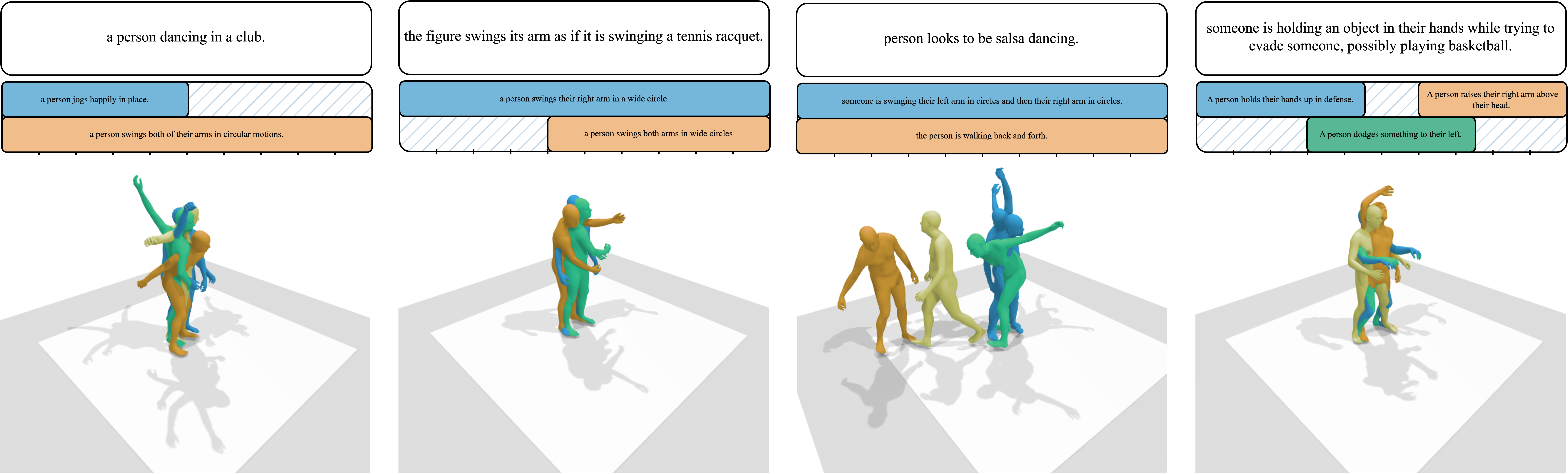}
    \caption{Examples of original textual annotations (upper white box), corresponding decomposed sub-movement annotations (middle box), and the generated motion conditioned on the sub-movement annotations (lower human figure).}
\label{fig:renders_examples_pipeline}
\vspace{-0.2cm}
\end{figure*}

\vspace{-0.3cm}
\subsubsection{Multi-annotations approach} \label{sec:multiannotations}
Our method allows combining multiple time intervals at inference time, each defined by a textual description and corresponding start and end times. 
HumanML3D~\cite{HumanML3D} and KitML~\cite{KIT} contain multiple textual annotations for each movement. This is because they were created using the Amazon Mechanical Turk interface, where each movement was presented to various users who provided textual annotations. On average, HumanML3D has 3.07 textual descriptions per movement, and KitML has 1.6.

To showcase our method's ability to combine more information than the state-of-the-art approaches, we use the various textual annotations of each movement in the dataset as the set of sub-movements necessary for generation. 
We set the start and end times for each annotation to match the overall movement's times. For this experiment, we do not use the GPT Decomposition module to obtain sub-movements; instead, we rely solely on the native information provided by the datasets.

We use for this experiment the same split employed by~\cite{human_motion_diffusion_model} and \cite{stmc} and the publicly available pretrained models to evaluate the combination capability across all movement classes, not just basic movements compared to complex ones. 
Results are shown in~\cref{tab:multiannotations_humanml_and_kitml}. 
For the HumanML3D dataset, we achieve better retrieval values, improved alignment of motion and text embeddings with the ground truth, and lower FID scores compared to generation conditioned on a single textual annotation. In contrast, for KITML, we see slightly better retrieval scores but slightly worse FID and M2M values. This is because KITML provides fewer annotations per dataset element compared to HumanML3D, leading to a smaller improvement in scores. These results indicate that our composition approach, which leverages all annotations together, produces better animations as the number of annotations to combine increases.

\vspace{-0.2cm}
\subsubsection{Ablation on the  hyperparameter $\boldsymbol{w}$}
An important aspect of our denoising combination process is the choice of the hyperparameter $w$, which regulates the ``\textbf{strength}'' of the combination of concurrent sub-movements. If $w$ is too small, none of the sub-movements are executed; if $w$ is too large, the signal becomes too strong, exceeding the diffusion process’s ability to render smooth animations, resulting in faster and more unnatural movements. 
The variation in results as $w$ changes for the split in Sec~\ref{splits} is shown in~\cref{tab:ablation_w_kitml_humanml_splitme}. It appears that for each dataset there is an optimal value of $w$ ($w=5$ for HumanML3D and $w=10$ for KITML). For lower values of $w$, we observe high FID, low retrieval scores, and low \textit{Transition} distances, indicating that the requested action is either not performed or only partially executed. In contrast, for values of $w$ larger than the optimal one, the \textit{Transition} distance increases, while the retrieval and FID scores worsen, as the increasingly faster animations become less realistic and they fall outside the real data distribution.

\subsection{Qualitative results}
We visualize some additional examples of the motions generated by \mname~in \cref{fig:renders_examples_pipeline}, using the decomposed sub-movement annotations as input. These results demonstrate that \mname~is capable of generating realistic motions for input prompts that have never been seen during training phases. 
Additional qualitative results and comparisons with motions generated by baseline methods are given as supplementary videos.

\subsection{Limitations}
Despite the results obtained and the advancement with respect to the current literature, our method still faces some limitations:
%
\emph{(i)} Its generation capability is closely tied to the GPT model used to decompose textual annotations into sub-movements. For our experiments, we used GPT-4o-mini, but newer versions, such as GPT-4o, would make the method more competitive;
\emph{(ii)} The decomposition ability of the GPT-decomposition module depends on the quantity and variety of actions present in the training set. If a dataset lacks sufficient basic actions, the decomposition of complex actions will be more coarse. This is evident in the experiments with the KITML dataset, where the improvement is lower compared to what is achieved with HumanML3D. This is due to KITML containing fewer elements, with a reduced number of basic action types;
\emph{(iii)} The ability of our method to generate realistic animations, despite being the result of multiple components, lies in the diffusion model's capacity to guide the generation towards the original data distribution. However, this also presents a limitation, particularly when composing two perfectly simultaneous sub-movements (with the same start and end times). In such cases, if no motion in the original dataset represents the two actions being performed together, it will be difficult to generate them correctly, as the diffusion model will tend to favor one component over the other, since individually it aligns better with the training set distribution that the model is trained to follow.

\section{Conclusions}
In this paper, we introduced a novel method for generating movements whose classes are not included in the training set, operating during inference without the need to train a new model. The method relies on a GPT-based module to decompose the input movement into known sub-movements, and a novel composition technique that accounts for the spatial and temporal overlap of movements. We split the datasets into basic actions, used for training, and complex actions, used as the test set, and demonstrated the effectiveness of our approach compared to text-only conditioned generation and state-of-the-art movement composition alternatives. Additionally, we tested the movement composition module on the task of synthesizing animations by considering multiple textual annotations for each dataset example. In this setting, we found that even without the GPT decomposition module, and by using the native information from the dataset, we were able to achieve better results than text-conditioned generation alone. 



{\small
\bibliographystyle{ieee_fullname}
\bibliography{main}
}

\begin{figure*}[h!]
    \centering    \includegraphics[width=\linewidth]{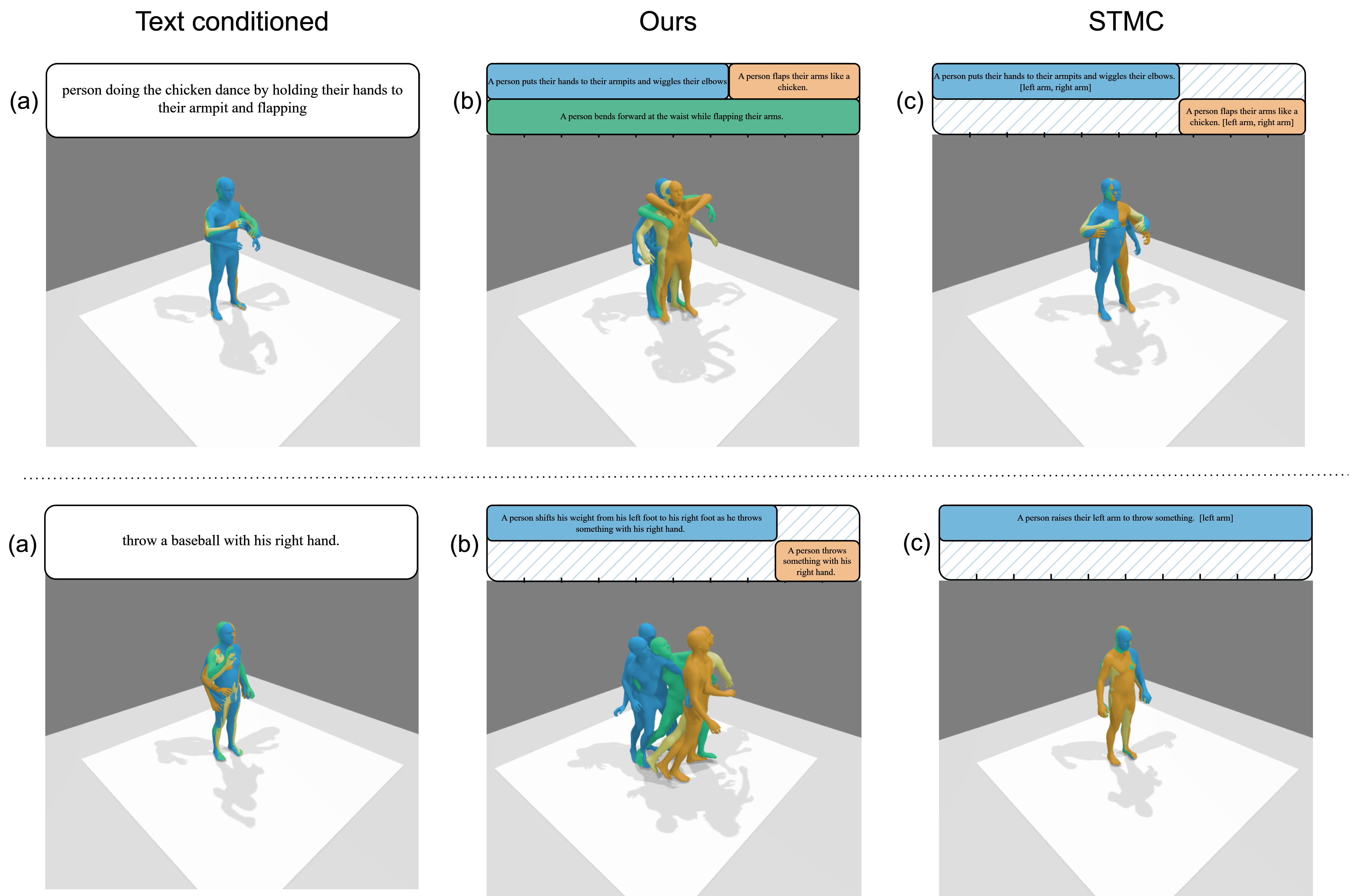}
    \caption{Examples of text and text-conditioned generation \textbf{(a)}, division into sub-motions and composition through MCD \textbf{(b)}, and division into sub-motions and composition through STMC \textbf{(c)}.}
\label{fig:renders_examples_more}
\end{figure*}

\newpage

\section*{Supplementary material}

\subsection*{Ablation on the hyperparameter $\boldsymbol{w}$}
Table \ref{tab:ablation_w_multiannotations} presents the ablation study for the hyperparameter $w$ on the splits used in \cite{human_motion_diffusion_model} and \cite{stmc} for the HumanML3D and KITML datasets, following the multi-annotation approach (Motion Composition) \ref{sec:multiannotations}. The trends in the metrics remain similar to those observed in the ablation study conducted on the base-complex action split used for MCD (GPT decomposition + Motion Composition) as presented in the main paper: the Transition value increases with $w$, indicating faster animations, while the other metrics exhibit an optimal range beyond which they progressively worsen.

\begin{table*}[h!]
\centering
\footnotesize
\caption{Ablation study varying the $w$ hyperparameter in the multi-annotations approach over the split used in \cite{human_motion_diffusion_model} andn \cite{stmc} for the datasets HumanML3D and KITML.}
\begin{tabular}{llccccccc}
\toprule
\textbf{Dataset} & \textbf{Experiment} & \textbf{R1 $\uparrow$} & \textbf{R3 $\uparrow$} & \textbf{R10 $\uparrow$} & \textbf{M2T S $\uparrow$} & \textbf{M2M S $\uparrow$} & \textbf{FID $\downarrow$} & \textbf{Trans $\downarrow$} \\
\midrule
\multirow{10}{*}{\textbf{HumanML3D}} 
    & GT & 6.259 & 15.487 & 33.548 & 0.762 & 1.0 & 0.0 & 1.36 \\
    \cmidrule(lr){2-9}
    & $w=1$    & 3.117   & 8.535   & 19.792  & 0.707  & 0.711  & 0.199  & 0.867 \\
    & $w=2$    & \textbf{4.503}   & 10.168  & 23.454  & 0.736  & 0.743  & 0.168  & 1.144 \\
    & $w=3$    & 4.379   & \textbf{10.218}  & \textbf{24.072}  & \textbf{0.74}   & \textbf{0.748}  & \textbf{0.167}  & \textbf{1.331} \\
    & $w=4$    & 3.859   & 9.55    & 23.726  & 0.737  & 0.746  & 0.173  & 1.448 \\
    & $w=5$    & 3.513   & 9.5     & 23.503  & 0.732  & 0.74   & 0.184  & 1.565 \\
    & $w=6$    & 3.835   & 9.179   & 22.563  & 0.726  & 0.734  & 0.197  & 1.683 \\
    & $w=7$    & 3.266   & 8.758   & 21.573  & 0.719  & 0.726  & 0.212  & 1.793 \\
    & $w=8$    & 3.513   & 8.684   & 19.619  & 0.714  & 0.72   & 0.224  & 1.886 \\
    & $w=9$    & 2.771   & 8.14    & 19.57   & 0.71   & 0.714  & 0.235  & 1.95  \\
    & $w=10$ & 2.771 & 7.472 & 18.605 & 0.704 & 0.709 & 0.248 & 2.025 \\
    & $w=15$ & 2.474 & 5.641 & 15.883 & 0.684 & 0.686  & 0.302 & 2.435 \\
    & $w=20$ & 2.35 & 5.468 & 13.063 & 0.669 & 0.669 & 0.346 & 2.721 \\
\midrule
\multirow{10}{*}{\textbf{KITML}} 
    & GT             & 7.888  & 19.847  & 41.858  & 0.803  & 1.0   & 0.0   & 1.534 \\
    \cmidrule(lr){2-9}
    & $w=1$ & 5.598  & 12.468  & 26.336  & 0.698  & 0.675  & 0.321  & 0.743 \\
    & $w=2$ & 7.506  & 15.903  & 32.697  & 0.739  & 0.714  & 0.234  & 0.886 \\
    & $w=3$ & 6.997  & 16.539  & 32.952  & 0.752  & 0.728  & 0.208  & 0.952 \\
    & $w=4$ & 6.997  & 17.43   & 34.733  & \textbf{0.755}  & \textbf{0.731}  & 0.195  & 1.013 \\
    & $w=5$ & \textbf{7.761}  & \textbf{18.193}  & \textbf{35.623}  & \textbf{0.755}  & \textbf{0.731}  & 0.192  & 1.069 \\
    & $w=6$ & 6.107  & 15.776  & 35.242  & 0.753  & \textbf{0.731}  & 0.192  & 1.159 \\
    & $w=7$ & 6.616  & 15.522  & 34.097  & 0.752  & 0.73   & 0.191  & 1.142 \\
    & $w=8$ & 6.743  & 15.903  & 34.606  & 0.751  & 0.73   & 0.189  & 1.175 \\
    & $w=9$ & 5.725  & 15.013  & 33.079  & 0.748  & 0.726  & \textbf{0.188}  & 1.204 \\
    & $w=10$ & 7.252 & 14.758  & 32.316  & 0.745  & 0.725  & 0.189  & 1.261 \\
    & $w=15$ & 6.361 & 13.74   & 29.262  & 0.731  & 0.714  & 0.2    & 1.391 \\
    & $w=20$ & 4.835 & 12.087  & 28.753  & 0.717  & 0.7    & 0.221  & \textbf{1.594} \\
\bottomrule
\end{tabular}
\label{tab:ablation_w_multiannotations}
\end{table*}

\section*{GPT Decomposition Prompts}

\subsection*{MCD prompt}
The instruction prompt used by our approach to initialize the GPT Decomposition module is shown in Listing \ref{code:prompt_ours}. The first lines specify the goal of the module: to decompose actions using known movements listed in the 'train' file, following examples provided in the 'examples' file. The answer is provided in a json format. To ensure GPT follows to the desired behavior, we added the following requirements: 
\begin{itemize}
    \item We emphasized the use of only words, actions, and verbs present in the file (line 9);
    \item We requested that the decomposition elements be as simple as possible to facilitate their combination (line~13);
    \item  We set a minimum action duration of 2 seconds, corresponding to the minimum movement length in the HumanML3D dataset (line 15); 
    \item We enforced standard temporal constraints, such as ensuring the start time is earlier than the end time and that at least one movement starts at zero (line 17-18);
    \item  We requested temporal overlap for consecutive actions whenever possible. Without this constraint, we observed that the character tends to return to a neutral position before performing each action. This is due to the structure of the dataset, where most actions include a reset phase after execution. (line 20)
\end{itemize}

\subsection*{STMC prompt}
As previously mentioned, for comparison with the STMC composition method, we required a different prompt for the GPT Decomposition module. This prompt is shown in Listing \ref{code:prompt_stmc}. Unlike our approach, it requests a different data format that includes the main body parts involved in each sub-motion (line 1). Additionally, we impose the following constraints:
\begin{itemize}
    \item Use only the allowed body parts: 'head', 'left arm', 'right arm', 'legs' and 'spine' (line~22);
    \item Ensure that no two temporal intervals use the same body parts (line 24);
    \item Ensure that there are no temporal intervals not assigned to any sub-movement (line 26).
\end{itemize}

For this method, partial overlap is not required to avoid resetting the character, as the time timelines are extended  for this reason during a preprocessing phase.

\begin{figure*}[!b]
\centering
\begin{lstlisting}[caption={Instructions used to initialize the GPT Decomposition module in our approach.}, label={code:prompt_ours}]
Provide valid JSON output. The output data schema should be like this: {"decomposition": [{"text": "string", "start": number, "end": number}, {"text": "string", "start": number, "end": number}, ...]}

I want to break down an action into a predetermined set of known actions. You have been provided with a list of known actions in the file called "texts_train".

You will be sent sentences in English one by one that describe the movement of a person with the start and end second of the movement.
The goal is to explain the input movement as if it had never been seen before, describing it as a combination of known movements found in the "texts_train" file.
Respond ONLY by breaking down the input action using the verbs and actions present in the "texts_train" file.

In the file "gpt_examples" there are some examples of decomposition. Use those as a reference.

YOU CANNOT USE MOVEMENT VERBS, ACTIONS, OR SPECIFIC NOUNS IF THEY ARE NOT IN THE FILE.

BREAK DOWN INTO SIMPLE SENTENCES, each focusing on a single body part, such as: "A person holds an object with his right hand". Avoid sentences composed of many clauses.

IF POSSIBLE DO NOT MAKE ANY ACTION LAST LESS THAN 2 SECONDS.

AT LEAST ONE OUTPUT ACTION MUST START FROM SECOND 0.
In each decomposition, the start second must be strictly less than the end second.

Try to ensure that the breakdowns have some temporal overlap of a few seconds.
\end{lstlisting}
\end{figure*}

\begin{figure*}[!b] 
\centering
\begin{lstlisting}[caption={Instructions used to initialize the GPT Decomposition module for STMC composition approach.}, label={code:prompt_stmc}]
Provide valid JSON output. The output data schema should be like this: {"decomposition": [{"text": "string", "start": number, "end": number, "body parts": list}, {"text": "string", "start": number, "end": number, "body parts": list}, ...]}

I want to break down an action into a predetermined set of known actions. You have been provided with a list of known actions in the file called "texts_train".

You will be sent sentences in English one by one that describe the movement of a person with the start and end second of the movement.
The goal is to explain the input movement as if it had never been seen before, describing it as a combination of known movements found in the "texts_train" file.
Respond ONLY by breaking down the input action using the verbs and actions present in the "texts_train" file.

In the file "gpt_examples" there are some examples of decomposition. Use those as a reference.

YOU CANNOT USE MOVEMENT VERBS, ACTIONS, OR SPECIFIC NOUNS IF THEY ARE NOT IN THE FILE.

BREAK DOWN INTO SIMPLE SENTENCES, each focusing on a single body part, such as: "A person holds an object with his right hand". Avoid sentences composed of many clauses.

IF POSSIBLE DO NOT MAKE ANY ACTION LAST LESS THAN 2 SECONDS.

AT LEAST ONE OUTPUT ACTION MUST START FROM SECOND 0.
In each decomposition, the start second must be strictly less than the end second.

For each sub-movement you create, indicate its textual description "text", the start time "start", the end time "end", and the involved "body parts."

Use ONLY the following body parts: "legs", "right arm", "left arm", "spine" and "head". DO NOT USE ANY OTHER WORDS EXCEPT THESE body parts.

CONSTRAINT: two sub-movements cannot be performed simultaneously or have any temporal overlap if they involve the same "body parts." For example, the following is NOT ACCEPTABLE: {"decomposition": [{"text": "a person raises the left arm", "start": 0, "end": 5.0, "body parts": ["left arm"]}, {"text": "a person throws a left punch", "start": 3.0, "end": 8.0, "body parts": ["left arm"]}]} because there is an overlap of "left arm" for the two sub-movements from second 3.0 to 5.0.

CONSTRAINT: Ensure that there is no time interval that is not assigned to any sub-movement. For example, the following is NOT ACCEPTABLE: {"decomposition": [{"text": "a person raises the left arm", "start": 0, "end": 2.0, "body parts": ["left arm"]}, {"text": "a person throws a left punch", "start": 4, "end": 6.0, "body parts": ["left arm"]}]} because there is a time interval from second 2.0 to 4.0 that is not assigned to any sub-movement.
\end{lstlisting}
\end{figure*}

\subsection*{Qualitative results}
In Figure \ref{fig:renders_examples_more}, we present additional qualitative results, highlighting the differences between text-conditioned generation, MCD, and STMC.
As shown, the decomposition phase produces two different sets of sub-movements for MCD and STMC, based on the different instructions provided. The decomposition obtained for STMC tends to be less precise than for MCD due to the greater difficulty in meeting the constraints and objectives of the task.

\end{document}